\pdfoutput=1

\documentclass[11pt]{article}

\usepackage[]{acl}

\usepackage{times}
\usepackage{inconsolata}  %
\usepackage{latexsym}
\usepackage{booktabs}
\usepackage{float}
\usepackage[T1]{fontenc}

\usepackage[utf8]{inputenc}

\usepackage{microtype}
\usepackage{graphicx}
\usepackage{tabularx}
\usepackage{multirow}
\usepackage{arydshln}
\usepackage{hyperref}
\usepackage{amsmath,amssymb,stmaryrd}
\usepackage[shortlabels]{enumitem}
\usepackage{listings}
\usepackage{fancyvrb}

\newcommand{\simple}{\textsc{Plain}}
\newcommand{\chain}{\textsc{MultiHop}}
\newcommand{\multi}{\textsc{MultiPreference}}
\newcommand{\multidomain}{\textsc{MultiDomain}}
\newcommand{\noinstruction}{\textsc{NoneApplicable}}
\newcommand{\override}{\textsc{Conflict}}
\newcommand{\static}{\textsc{ICL}}
\newcommand{\dynamic}{\textsc{ICL-Dynamic}}
\newcommand{\multipass}{\textsc{Multi-Pass}}
\newcommand{\oracle}{\textsc{Oracle}}
\newcommand{\embedding}{\textsc{Contriever}}
\newcommand{\direct}{\textsc{Direct}}
\newcommand{\joined}{\textsc{Select-And-Interpret}}
\newcommand{\bm}{\textsc{BM25}}

\newcommand{\scenario}{reasoning type}
\newcommand{\scenarios}{reasoning types}
\newcommand{\scenarioCap}{Reasoning Type}
\newcommand{\scenariosCap}{Reasoning Types}

\newif\ifshowchanges
 
\ifshowchanges
  \newcommand{\newremove}[1]{{\color{blue} \sout{#1}}}
  \newcommand{\newreplace}[2]{{\color{blue} \sout{#1}}{\color{red} #2}}
  \newcommand{\new}[1]{{\color{red} #1}}
  \newcommand{\newr}[1]{{\color{green} #1}}
\else
  \newcommand{\newremove}[1]{}
  \newcommand{\newreplace}[2]{{}{#2}}
  \newcommand{\new}[1]{{#1}}
  \newcommand{\newr}[1]{{#1}}
\fi

\newcommand{\dataname}{\textsc{NLSI}}

\title{Interpreting User Requests in the Context of \\
Natural Language Standing Instructions}

\author{Nikita Moghe\textsuperscript{$\gamma$}\thanks{{ } Work done while interning at Microsoft} \quad Patrick Xia\textsuperscript{$\Phi$} \quad Jacob Andreas\textsuperscript{$\Phi$} \\\bf{Jason Eisner\textsuperscript{$\Phi$} \quad  Benjamin Van Durme\textsuperscript{$\Phi$} \quad Harsh Jhamtani\textsuperscript{$\Phi$}}
\\
\textsuperscript{$\gamma$}School of Informatics, University of Edinburgh \\
\textsuperscript{$\Phi$}Microsoft Semantic Machines\\
\normalsize \texttt{nikita.moghe@ed.ac.uk hjhamtani@microsoft.com}
}

\begin{document}
\maketitle

\begin{abstract}

Users of natural language interfaces, frequently powered by Large Language Models (LLMs), must often repeat their full set of preferences each time they make a similar  request. 
We describe an approach to LLM-based dialogue modeling in which persistent user constraints and preferences -- collectively termed \emph{standing instructions} -- are provided as additional context for such interfaces. For example, when a user states \emph{I'm hungry}, a  previously expressed preference for Persian food can be automatically added to the LLM prompt, influencing the search for relevant restaurants.
We develop \dataname{}, a language-to-program dataset consisting of over 2.4K English dialogues spanning 17 domains, in which each dialogue is paired with a user profile (a set of user-specific standing instructions) and corresponding structured representations (a sequence of API calls). A key challenge in \dataname{} is to identify which subset of the standing instructions is applicable to a given dialogue. \dataname{} contains diverse phenomena, from simple preferences to interdependent instructions such as triggering a hotel search whenever the user is booking tickets to an event. 
We conduct experiments on \dataname{} using prompting with large language models and various retrieval approaches, achieving a maximum of 46\% exact match on API prediction.  
Our results demonstrate the challenges in identifying the relevant standing instructions and their interpretation into API calls\footnote{Code: \url{https://github.com/nikitacs16/nlsi}\\ Data: \url{https://huggingface.co/datasets/nikitam/nlsi}}.

\end{abstract}

\section{Introduction}
\begin{figure}[t]
    \centering
    \includegraphics[width=.49\textwidth,clip]{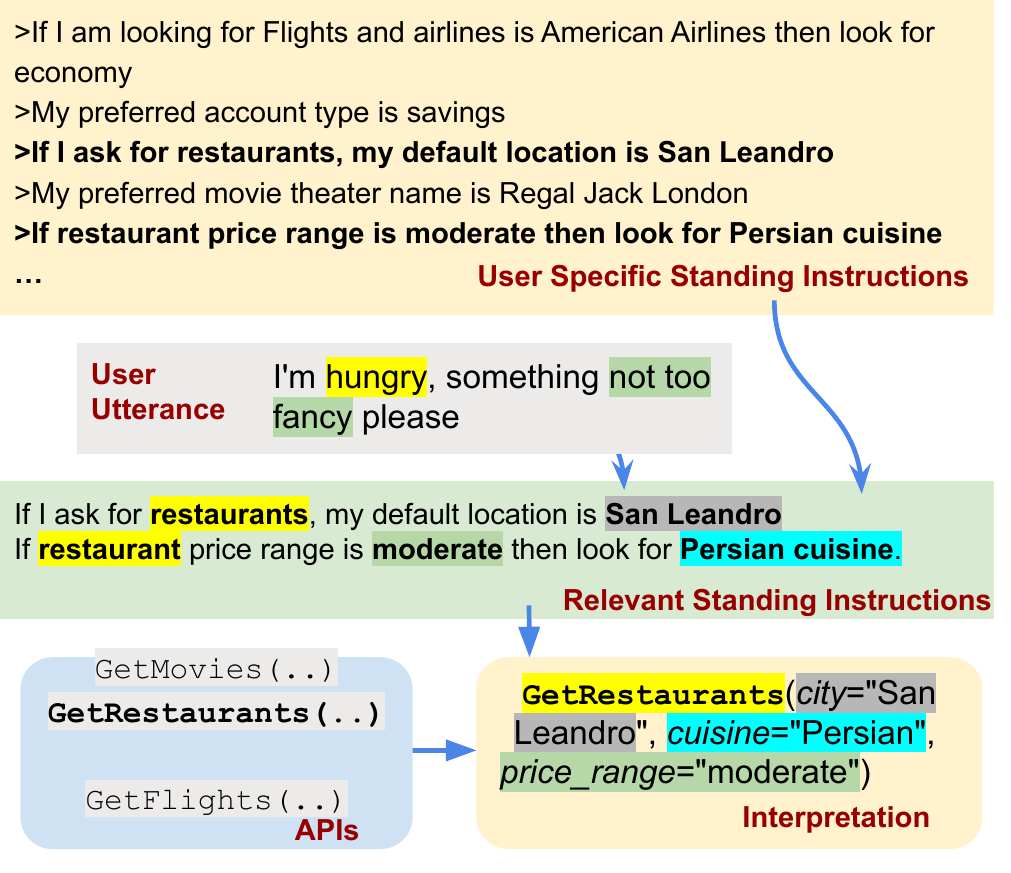}
    \caption{Parsing an utterance into a structured output, in the presence of a \emph{user-specific} set of \emph{standing instructions}. A model for the task needs to identify (explicitly or implicitly) the subset of instructions applicable to the utterance and interpret the utterance into API calls. 
    }
    \label{fig:nlsi}
\end{figure}

Large language models (LLMs) such as GPT-3 \cite{brown2020language}, GPT-4 \cite{openai2023gpt4}, and LLaMa 2 \cite{touvron2023llama} are increasingly being used with tools and APIs \cite{schick2023toolformer, qin2023toolllm} to provide additional functionality to users. For example, ChatGPT allows several external plugins such as OpenTable for searching and reserving restaurants or booking travel through Expedia.\footnote{\url{https://openai.com/blog/chatgpt-plugins}}
\newremove{or solving math problems with Wolfram}\newremove{As the same interface provides multiple services,}These applications must learn to identify which service the user is seeking while respecting preferences across diverse domains that are unique to each user. Understanding such preferences can aid in personalising the user experience by providing tailored responses, increased accuracy in recommendations and saving user time.
However, in most cases, users must verbalise their preferences in detail during the interaction, including for repeated requests. %

Past work has explored learning preferences from user-system interactions over time \citep{micarelli2007personalized,salemi2023lamp}. These preferences can be hard to learn while also requiring significant amounts of training data. Further, these learnt preferences are \textit{implicit} and usually cannot be interpreted or edited by the user.

We propose incorporating personalised \textit{standing instructions} explicitly as additional context while interpreting a user's requests. Standing instructions are user-provided %
natural language statements to change or prescribe system behaviour under certain circumstances. For example, in Fig.~\ref{fig:nlsi}, the user wishes to look for some nearby restaurants. In the absence of standing instructions, \newreplace{the system will likely proceed by asking where the user is located followed by their favourite cuisine}{the user might have to interact for multiple turns with the system to arrive at their preferred restaurant cuisine and location}. By looking up the relevant standing instructions for restaurants, the system can directly search for \textit{Persian restaurants} in \textit{San Leandro}, saving the user's time as well as providing customised/localised recommendations.
Explicit natural language instructions are also both controllable and interpretable. A user can inspect and edit their standing instructions, especially for preferences that change over time.  Further, the generated outputs can be directly linked to the relevant standing instructions, improving the user's trust in the system \citep{liu2023trustworthy}. 

Our work is related to \citet{gupta2022dialguide}, which conditions a dialogue model's response on a set of developer guidelines. Their work focuses on controlling response generation in open-domain dialogue systems with a focus on reducing toxicity and enhancing safety. More recently, 
\newr{commercial LLM providers have introduced System Prompts\footnote{\url{https://docs.anthropic.com/claude/docs/how-to-use-system-prompts}}/Custom Instructions\footnote{\url{https://openai.com/blog/custom-instructions-for-chatgpt}}/Preamble\footnote{\url{https://txt.cohere.com/chatbot-chat-endpoint/}} which have an option to include guidelines at the beginning of every conversation to improve response generation.}
However, not much is known about how it operates, and no evaluations of its usage have been documented, \newr{especially in the task-oriented setting}.

This work makes the following contributions: 
\newreplace{\begin{enumerate}
    \item We systematically study the incorporation of standing instructions in a task-oriented setup. 
We develop and introduce \textbf{\dataname{}} (Natural Language Standing Instructions), an English-language dataset in which every example consists of a conversation between a user and a dialogue agent, accompanied by a collection of standing instructions (a \textit{user profile}) and a sequence of API calls that can be inferred from the context of the conversation and the relevant standing instructions.
\item We investigate six \scenarios{} for using standing instructions that range from the inclusion of simple attributes to more complex situations such as the user making new choices over existing preferences, the user proposing multiple preferences, \textit{etc.} 
These \scenarios{} introduce challenges pertaining to
\begin{enumerate}[nosep]
\item identifying which subset of standing instructions is relevant for the conversational context. Whether an instruction is relevant or not is a function of user utterance and past turns, and requires multihop and cross-domain reasoning.
\item effectively incorporating standing instructions while inferring the structured API calls and their attributes. This may require joint reasoning over the dialogue and the relevant standing instructions, as well as dealing with any conflicts between user utterance and instructions.
\end{enumerate}
\item  We use this dataset to benchmark combination of methods involving the selection and interpretation of standing instructions. We observe that LLM-based methods are far from perfect, raising new challenges in retrieval, reasoning, and semantic parsing. 

\end{enumerate}}
{
\textbf{(i)} We systematically study the incorporation of standing instructions in a task-oriented setup. 
We develop and introduce \textbf{\dataname{}} (Natural Language Standing Instructions), an English-language dataset in which every example consists of a conversation between the user and a dialogue agent, accompanied by a collection of standing instructions (\textit{user profile}) and a sequence of API calls reflecting user intents. %
\textbf{(ii)} We investigate six \scenarios{} for using standing instructions that range from a single instruction for a specific attribute to more complex situations such as the user proposing multiple preferences for same aspect, \textit{etc.} 
These \scenarios{} introduce challenges pertaining to
subset selection of relevant standing instructions and then inferring the structured API calls and their arguments. These include instructions that specify a single preference to more complex ones that involve multi-hop, cross-domain, and conflict reasoning.
\textbf{(iii)} We use this dataset to benchmark a variety of methods involving the selection and interpretation of user utterances in the presence of standing instructions. We observe that our LLM-based methods are far from perfect, raising new challenges in retrieval, reasoning, and semantic parsing. 
}

\section{Task Overview}
\label{sec:overview}

We are interested in translating a user utterance into a sequence of API calls in the context of user-specific standing instructions (Figure \ref{fig:nlsi}). 
Consider a conversational context $x$, which consists of dialogue history between the user and the system (if any) and the user's current utterance. We assume a user profile $u$ consisting of a sequence of natural language instructions $u_1,u_2,...u_M$. 
In this setting, instruction following consists of a \textit{selection} task (which obtains a set of standing instructions $z$ from the user profile $u$ that are relevant to $x$)
followed by an \textit{interpretation} task (which predicts API calls $y$ based on the conversational context and the relevant subset of standing instructions $z$).
We assume access to a schema $s$ that lists the valid API method names and their keyword arguments (slots).
Formally, an agent of this kind is described by a generative model:
\begin{gather*}
z \sim p(\cdot \mid x,u) \\
y \sim p(\cdot \mid x,z,s)
\end{gather*}

\newcolumntype{K}{>{\arraybackslash}m{1.5cm}}
\newcolumntype{L}{>{\arraybackslash}m{3.9cm}}
\newcolumntype{M}{>{\arraybackslash}m{4.2cm}}

\begin{table*}[]
\scriptsize
\begin{tabular}{KMMM}
\toprule
 & \simple{} & \chain{} & \multi{} \\ \midrule
Relevant \newline Standing \newline Instructions \newline ($z$) & >\textcolor{red}{I always go to Santa Rosa if I'm looking for Movies.} \newline >\textcolor{red}{I like fantasy movies the best.} & >\textcolor{red}{If I'm looking for a flight, American Airlines is my go-to.}\newline>\textcolor{red}{If I'm flying American Airlines, check for Economy seating class.} & >\textcolor{red}{If I ask for Events, my preferred event type is Music.} \newline
>\textcolor{red}{When the event type is Music, search for Blues as the category.} \newline
>\textcolor{red}{Search for the event name Greensky Bluegrass if the category is Blues.}
\newline >\textcolor{red} {If I ask for Events, my preferred event type is Sports.}
\\\\

Conversation ($x$) & \textcolor{blue}{\emph{User}: I want to go out to watch a movie, please help me find a good one.} &  \textcolor{blue}{\emph{User:} Can you get on and get me a round trip ticket?}
\newline \emph{Agent:} Where will you go? Where are you coming from?\newline \textcolor{blue}{\emph{User:} I'm going to SFO from New York City.}
& \textcolor{blue}{\emph{User:} My schedule is free today and I plan to go to an event in Seattle, WA. I want to look for events in that area.} \\

API calls ($y$) & \begin{lstlisting}[basicstyle=\scriptsize\ttfamily, breaklines=true]
GetMovies(
  genre="fantasy",
  location="Santa Rosa")
\end{lstlisting} & 
\begin{lstlisting}[basicstyle=\scriptsize\ttfamily, breaklines=true]
GetFlights(
  destination="SFO",
  origin="New york",
  airlines="American Airlines",
  seating_class="Economy")
\end{lstlisting}
& %
\begin{lstlisting}[basicstyle=\scriptsize\ttfamily, breaklines=true]
GetEvents(
  city="Seattle, WA",
  event_type="Music",
  category="Blues",
  event_name="Greensky Bluegrass")
GetEvents(
  city="Seattle, WA",
  event_type="Sports")
\end{lstlisting}

\\
\bottomrule
\end{tabular}
\caption{Some examples from \dataname{}. 
User profile is not shown %
for brevity.
(1) In \simple{}, %
the instructions usually represent a domain matching problem. 
(2) In \chain{}, note that the seating class attribute \textit{Economy} is dependent on choosing the instruction with \textit{American Airlines}. 
(3) For the example for \multi{}, as there are two preferences for the same attribute \texttt{event\_type}, there are two separate API calls. Further, the API call with \texttt{event\_type} \emph{Music} has additional attributes. Additional examples are provided in the Appendix.
}
\label{tab:combined_examples}
\end{table*}
\section{Dataset: \dataname{}}
\label{sec:dataset_creation}
Existing related datasets have focused on generating safer responses in open-domain dialogue via natural language guidelines \citep{gupta2022dialguide} or looked at personalised text generation by conditioning on a set of past user-written documents like emails or reviews \citep{salemi2023lamp}. Similarly, \citet{madaan2022memory} improved response generation on user feedback on past conversations to assist new users on tasks such as ethical reasoning and word scrambling. Within task-oriented dialogue, the works in \citet{joshi2017personalization, 10.3389/frobt.2021.676814} focus on personalisation with a small set (<5) of preferences. 
Due to the lack of comprehensive datasets that study the use of natural language standing instructions in a language-to-program setup, we created \dataname{}. Our dataset covers multiple domains like airline booking or finding events. Each domain has an associated API.

\subsection{\scenariosCap{}}
In the context of standing instructions, various types of reasoning might be needed to predict API calls. Following a single standing instruction may be easier than composing and reasoning over several instructions. Furthermore, reasoning across several instructions in the same domain, like booking hotels, may be easier than across domains. Thus, to enable comparisons at different difficulties, we designated six \scenarios{} for \dataname{}. While these are not exhaustive, they allow us to systematically study a range of situations ranging from simple domain matching to more complex reasoning (examples in Table~\ref{tab:combined_examples}):

\paragraph{\noinstruction{}} For these examples, no standing instructions from the user profile are required for interpreting the user's utterance ($z = \emptyset$).

\paragraph{\simple{}} These examples use the standing instructions directly: each argument can be predicted from a single standing instruction. All the relevant standing instructions, $z$, belong to the same domain.
\newcolumntype{K}{>{\arraybackslash}m{3cm}}
\newcolumntype{M}{>{\arraybackslash}m{6.5cm}}
\newcolumntype{L}{>{\arraybackslash}m{4.5cm}}

\begin{table*}[]
\scriptsize
\centering
\begin{tabular}{MKL}
\toprule
SGD &
  Action &
  \dataname{} \\ \midrule
User: Can you get on and get me a round trip ticket? &
  use as dialogue & \textbf{Dialogue:}
  User: Can you get on and get me a round trip ticket? \\ 
  
Agent: Where will you go? Where you coming from? &
  use as dialogue &
  Agent: Where will you go? Where you coming from? \\
User:I'm going to \textcolor{blue}{SFO} from \textcolor{blue}{New York City}. &
  \begin{tabular}[c]{@{}l@{}}use as dialogue,\\ use as parameters\end{tabular} &
  User: I'm going to SFO from New York City. \\ 
Agent: When are you leaving? When will you return? &
  discard & \textbf{Standing Instructions:} \\

User: I need to get back on the 14th. I really insist on getting \textcolor{blue}{American Airlines} tickets. I have mile advntage with them. I'm taking off on Sunday this week. &
  convert to standing instruction &   \multirow{2}{*}{\begin{tabular}[c]{@{}l@{}}If I’m looking for a flight, American Airlines \\ is my go-to \end{tabular}} 
   \\
Agent: You're in luck, there's an American Airlines flight that takes off at 8:50 pm. You'll return leaving at 8:55 pm. You'll only pay \$203 for everything. &
  discard  &
   \\
User: Ok, just make sure I get the best \textcolor{blue}{economy} deal &
  \begin{tabular}[c]{@{}l@{}}convert to standing instruction,\\ dependent on the previous one\end{tabular} &  If I’m flying American Airlines, check for
Economy seating class
   \\\\
Agent: Ok to be clear: 1 ticket from New York going to San Francisco on American Airlines at 8:50 pm on March 3rd, economy. You'll return boarding at 8:55 pm on March 14th. &
  discard this and future turns & \textbf{API Call}: 
\begin{lstlisting}[basicstyle=\scriptsize\ttfamily, breaklines=true]
GetFlights(
  destination="SFO",
  origin="New York City",
  airlines="American Airlines",
  seating_class="economy")
\end{lstlisting} \\
  
 \bottomrule
\end{tabular}
\caption{Converting an example from SGD dataset \citep{rastogi2020towards} into \dataname{} format. We show a per utterance decision process to obtain the dialogue, standing instructions, and parameters for the API call. We exclude parameters that cannot be converted into standing instructions. We exclude utterances not relevant to the creation of standing instructions. }
\label{tab:sgd-nlsi-conversion}
\end{table*}

\paragraph{\chain{}} These examples contain at least one standing instruction in $z$ that is relevant to the dialogue $x$ only due to the
presence of another standing instruction in $z$. These are of the form ``if A then B'' and ``if B then C'', where A, B, and C are slot names from the same domain. For example, choosing seating\_class as \textit{economy} is dependent on choosing airlines as \textit{American Airlines}. %
These examples test multi-hop reasoning abilities of the model. %
\paragraph{\multidomain{}} These examples are like \chain{} except that there is at least one relevant instruction in $z$  that links two domains. 
These example types typically involve triggering API(s) from an additional domain while being consistent on any shared arguments such as location. For example, the user might request searching for Hotels when looking for places to visit (Travel).
These example types require the identification of standing instructions relevant to either domain as well as sharing any common attributes, like location or date, across the domains. 
These examples challenge multi-domain understanding in addition to multi-hop reasoning. %

\paragraph{\multi{}} These examples contain standing instructions catering towards multiple preferences for the same attribute. %
The interpretation task for such examples requires placing multiple API calls respecting the different constraints (\textit{Music} or \textit{Sports} when picking an event type).    
\paragraph{\override{}} These examples include instructions in the profile $u$ that conflict with the information in the user utterance in the dialogue $x$. %
The model should gracefully handle such situations and give preference to the user's request.

Examples can contain standing instructions demonstrating multiple \scenarios{}. In NLSI, we associate each example with a single type as based on the above ordering - a type occurring later in the above ordering gets precedence.

\subsection{Dataset Creation}
\label{sec:instruction_types}
We constructed \dataname{} by extending Schema Guided Dataset  \citep[SGD,][]{rastogi2020towards}. SGD consists of multi-turn conversations across 20 domains like airlines or restaurants. 
We chose SGD because the dialogues in that dataset include natural and rich conversations 
and the accompanying annotations make it possible to construct the ground truth API labels.
The process outlined below intends to repurpose an existing dataset for studying the selection and interpretation tasks. %
In a real-world setting, a user might provide explicit preferences through another interface, or else such preferences would be inferred from the user's continuous interaction with the system. \new{We briefly discuss the dataset creation below and provide details in Appendix~\ref{app:data}}.

\paragraph{Extracting standing instructions:}  

We first identified which slots within the SGD schema can be translated into standing instructions based on the slot descriptions provided in the original dataset. \new{For example,  \texttt{theatre\_name} is inclined to be a persistent user preference unlike \texttt{movie\_title} or \texttt{date} which are likely to change with every interaction.}
\newremove{For example, while booking movie tickets, \texttt{theatre\_name} is inclined to be a persistent user preference, hence it can be part of a standing instruction. In contrast, \texttt{movie\_title} or \texttt{date} of booking the movie ticket should not be converted to standing instructions, as these are likely to change every time the user interacts with the system.}

Each conversation in SGD originated from a sequence of actions that a user or agent should take \new{alternately}. \newreplace{For example, Greet() $\rightarrow$ Inform(location) $\rightarrow$ Request(cuisine) $\rightarrow$ Inform(date) $\rightarrow$ Offer(restaurant\_name). }{For example, the second conversation in  Table~\ref{tab:combined_examples} was based on a template sequence like \texttt{Inform(airline\_ticket)} $\rightarrow$ \texttt{Request(origin, dest)} $\rightarrow$ \texttt{Inform(origin, dest)} $\rightarrow$ \texttt{Offer(airlines)} $\rightarrow$ \texttt{Confirm(airlines)}, \texttt{Request(seating\_class)}.}
\newr{These sequences were then specialized by binding the variables, and the resulting sequence was written as a dialogue by a crowd worker that constituted this SGD example.} We reverse-engineer the original SGD creation process to construct the standing instructions for NLSI. 

To convert an SGD dialogue to an NLSI dialogue with standing instructions, we retained the first one or three turns as the conversational context $x$, and converted the remaining turns into the relevant standing instructions $z$. See an illustration in Table~\ref{tab:sgd-nlsi-conversion}.
\newreplace{We ignored any turns that could not be converted into instructions. For example, the second NLSI example in Table~\ref{tab:combined_examples} was derived from an SGD dialogue that had originally continued with natural language turns that specified \texttt{airlines=``American
Airlines'',
seating\_class=``Economy''}.}{Continuing our example, the natural language turns that specified \texttt{airlines=``American
Airlines'',
seating\_class=``Economy''} were converted to standing instructions.  We excluded information from any turns that could not be converted into a standing instruction - see the sixth utterance in the table}.

\newr{We start with templated instructions for different scenarios in an if-then format akin to the work in \citet{gupta2022dialguide}. To convert these templated instructions into natural language, we use GPT-3 to paraphrase the templated instructions and obtain diverse instructions.
We list the prompts to obtain these paraphrases in Appendix~\ref{app:data}.}
\newremove{Those remaining turns are not needed to predict $y$ from $x$ provided that the standing instructions $z$ can be detected as relevant to that prediction.} 

\newremove{Revisiting the example SGD series of rules mentioned earlier, the dialogue corresponding to Greet() and Inform(location) becomes part of the conversation context $x$ and Request(cuisine) and Offer(restaurant\_name) become the standing instruction. The instruction is templated as  ``If I ask about \textit{Restaurants}, my preferred \textit{cuisine} is \textit{Italian}''.  As \textit{date} is a non-instructional slot, we exclude it.  }

\newremove{Additional details on how examples for various \scenarios{} are constructed are discussed in Appendix \ref{app:data}}

\paragraph{Forming user profiles:} The above process provides us with the \emph{relevant} standing instructions $z$ for the given example from SGD, but these are only part of the full user profile $u$. A user will have additional preferences that are not relevant to the given example.
To emulate this, for the given example, we create $u$ by augmenting $z$ with $M$ randomly sampled instructions from other examples.
\newremove{\footnote{We drew $M$ uniformly from the range $[3,12]$. %
In particular, we drew the distractor instructions before splitting the dataset into train/dev/test, so training examples were constructed with some distractors sourced from the test set.  
Given this dataset, however, our experiments followed the usual protocol of holding out the test set while constructing our systems.}} These ``distractor'' instructions are sampled from domains unrelated to the current domain(s). %

\newremove{ \paragraph{Post-processing of instructions:}
We also included several rounds of pre-checks and post-processing on the dataset to remove undesirable or unrealistic situations that arise either through the noise in the base dataset or our extraction process like domain mismatch (``Play music'' followed by ``Book me a bus ticket''). To make the standing instructions more natural and diverse, we paraphrased the templates using LLMs (GPT-3.5).}

\paragraph{API calls:}
The outputs of the interpretation task are API calls $y$, in line with the recent works of integrating LLMs with tools and plugins \citep{schick2023toolformer, qin2023toolllm}.  The API calls are of the format \texttt{GetDomain(slot\_1=value\_1, slot\_2=value\_2)}. The argument names and values are derived from annotations in the SGD examples, which are either mentioned in the user's utterance or inferred in the standing instructions.

\paragraph{Dataset Statistics:}
We construct a balanced test set based on the different \scenarios{}: 340 per \scenario{}, leading to a total of 2040 examples across 17 domains. The train set contains at most 10 examples per domain with a minimum of five examples per \scenario{}, for a total of 150 examples. The remaining examples form the development set (251). There are 10.4 $\pm$ 3.0 instructions in a user profile (min: 3, max: 22) and there are 2.1 $\pm$1.7 relevant standing instructions per example in the dataset (min: 0, max: 10). There are 17 function calls corresponding to the 17 domains.

\begin{figure*}
    \centering
    \includegraphics[scale=0.5]{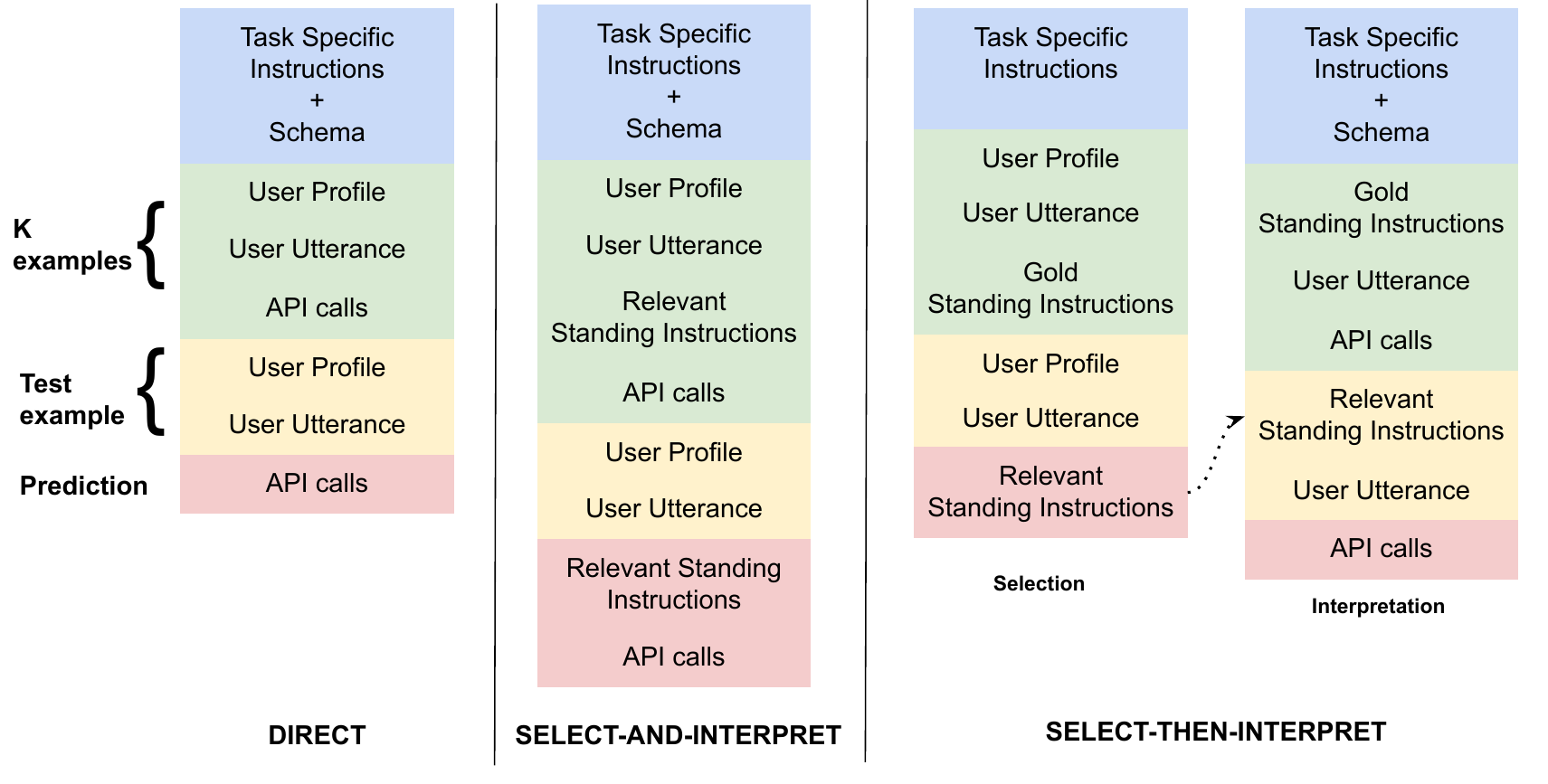}
    \caption{Illustration of different prompting methods.
    The blocks in red are the expected output generation and every other block is part of the input. The green bits are repeated $K$ times, providing $K$ demonstrations for in-context learning. $\direct{}$ Interpretation conditions the generation of API calls on the user profile and user utterance.
    $\joined{}$ requires the generation of the appropriate standing instructions based on user profile and user utterance followed by API generation.
    $\textsc{Select-Then-Interpret}$ receives the predicted standing instructions from a separate Selection Model (see left) in addition to the user utterance and then generates the API calls. The selection step only generates the standing instructions based on the user profile and the user utterance. }
    \label{fig:baselines}
\end{figure*}

\section{Methods}
\label{sec:methods}

Given the recent success of using LLMs to generate outputs in structured prediction tasks \citep{roy2022benchclamp, schick2023toolformer, heck-etal-2023-chatgpt}, we use an LLM-based method to interpret a user utterance into a structured API call.
We use in-context learning \citep{dong2023survey} by providing $K$ demonstrations, where $K$ is tuned on the dev set. These demonstrations are obtained by retrieving examples from the training set that are most similar to the current dialogue of the test example using the BM25 similarity measure \citep{Robertson1994OkapiAT} as in \citet{rubin-etal-2022-learning, roy2022benchclamp}. The examples are arranged in a best-first order. 
We describe the different paradigms (Fig.~\ref{fig:baselines}) used for the interpretation task by selecting the instructions implicitly (\direct{} Interpretation), jointly (\joined{}) or explicitly (\textsc{Select-Then-Interpret}).

\subsection{Direct Interpretation}

In the \textbf{\direct{}} method, we do not have any explicit selection of standing instructions from the user profile, and directly interpret the dialogue context into API calls.
The input to the LLM (Fig.~\ref{fig:baselines}) consists of (i) instructions about the interpretation task including the information about using standing instructions, (ii) the schema of the dataset (list of functions and arguments that can be used when generating API calls) $s$, (iii) user profile $u$, (iv) user's dialogue $x$, and (v) API calls $y$. Of these, (iii)-(v) are repeated for every demonstration example and the test example only consists of the user profile and the dialogue. We also include the list of categorical slots and their categories as well as a list of boolean slots while describing the schema. \newr{This method is similar to the commercial usage of System Prompts.}
This setup allows us to evaluate the ability of implicit selection of the relevant standing instructions for the interpretation task.

\subsection{Joint Selection and Interpretation}
Inspired by the effectiveness of techniques like \textit{Chain-of-Thought} prompting \citep{Wei2023CoT} across several tasks \citep{chu2023survey}, we also treat the direct interpretation task with a two-step approach: generate the relevant standing instructions $z\subseteq u$ and then generate the corresponding API calls $y$. \newremove{In addition to potentially improving the accuracy,} Such explicit selection can enhance the transparency of the system by exposing the relevant subset of instructions to the user \citep{liu2023trustworthy}.
To implement the method, the input prompt to the LLM is modified such that the demonstrations include the set of all standing instructions $u$, the relevant standing instructions $z$, and then the API calls $y$ (Fig.~\ref{fig:baselines}). We refer to this method as \textbf{\joined{}}.

\subsection{Selection Then Interpretation}
\label{sec:select-then-interpret}
Here we treat selection and interpretation with two separate models.%
\newremove{The selection model is not limited to an LLM-based approach.} The interpretation model is similar to the one described for \direct{}, except that instead of user profile, the relevant standing instructions are used directly. By decoupling the selection task from the interpretation task, we can explore popular methods of information retrieval for selection. As the user profile size increases, and the instructions no longer fit into the prompt, a separate selection step can be convenient.
We now describe various approaches for the selection step. 

\paragraph{\oracle{}:} The selection step simply returns the true $z$.
This setup measures the standalone performance of the interpretation task when given the correct standing instructions.

\paragraph{\bm{}:} The selection step sets $z$ to the $N$ instructions from the user profile $u$ that are most similar to the dialogue $x$ \new{using BM25 \citep{Robertson1994OkapiAT}}, where $N$ is tuned on the dev set.  To compute the corpus statistics \newreplace{used by BM25 to define similarity}{for BM25}, each instruction in $u$ is considered a document, and as is each standing instruction from the train examples.   

\paragraph{\embedding{}:} 
As above, replace BM25 with cosine similarity.
The dialogue $x$ and each standing instruction in $u$ is embedded into $\mathbb{R}^{768}$ with a pretrained sentence encoder, \textsc{Contriever} \citep{contriever}.  %
Both \bm{} and \embedding{} have been used as baselines in similar past work \cite{gupta2022dialguide, salemi2023lamp}.

\paragraph{\static{}:} We 
also experiment with using LLMs for the selection task. %
The fixed input prompt to the LLM consists of instructions for the selection task, followed by exactly six demonstrations, each consisting of a dialogue $x$, user profile $u$, and relevant standing instructions $z$ and then the  test example (see Fig.~\ref{fig:baselines}, Selection). 
We randomly sampled the six demonstrations from the training set, one per \scenario{}, and used the same demonstrations for all the test examples.

\paragraph{\dynamic{}:}
Similar to \static{}, except that  
now $K$ demonstrations are dynamically retrieved from the train split by using the ones that are similar to the dialogue in the current example through BM25.

\paragraph{\multipass{}:} 
In our preliminary experiments with \newremove{the previous} LLM-based selection methods, we observed that the LLMs consistently missed a subset of relevant instructions in the \chain{} and \multidomain{} \scenarios{}. %
We propose running the selection step multiple times to add these missing instructions. We use the standing instructions selected in the first pass of the selection process from \static{} as part of the prompt to perform a new selection step. We instruct the model to find the standing instructions that are missing from the current selection set.
Though the process can be iterated across multiple steps, %
we found the best results with only one additional round of selection.

\begin{table*}[t]
\centering
\small
\resizebox{\textwidth}{!}{
\begin{tabular}{@{}l|cccc|cccc|cccc@{}}
\toprule
\bf  & \multicolumn{4}{c}{\bf GPT-3.5} & \multicolumn{4}{c}{\bf GPT-4} & \multicolumn{4}{c}{\bf LLaMA 2 (7B)}\\
\bf Method & \multicolumn{2}{c}{\bf Selection} & \multicolumn{2}{c}{\bf Interpretation} & \multicolumn{2}{c}{\bf Selection} & \multicolumn{2}{c}{\bf Interpretation} & \multicolumn{2}{c}{\bf Selection} & \multicolumn{2}{c}{\bf Interpretation} \\

& EM$\uparrow$ &  F1$\uparrow$ & EM$\uparrow$ & F1$\uparrow$ & EM$\uparrow$ &  F1$\uparrow$ & EM$\uparrow$ & F1$\uparrow$ & EM$\uparrow$ & F1$\uparrow$ & EM$\uparrow$ &  F1$\uparrow$ \\  \midrule
\direct{} & N/A & N/A & {\bf 32.0} & {\bf 66.4} & N/A & N/A & 42.0 & 67.9 & N/A & N/A & {\bf15.1} & {\bf 47.8}  \\
\joined{} & 25.9 & 50.3 & 28.0 & 65.9 & 46.5 & 67.6 & 40.2 & 73.2 & 12.0   & 26.2 & 15.0       & 47.7          \\ \hline
\textsc{Select-Then-Interpret} & & & & & & & & & & & \\
\quad \bm{} & 17.3 & 19.3 & 11.2 & 39.7 & 17.3 & 19.3 & 11.8 & 40.8 & {\bf 17.3} & 19.3 & 7.8 & 30.9 \\
\quad \embedding{} & 14.6 & 51.5 & 17.2 & 57.5 & 14.6 & 51.5 & 25.4 & 62.7 &  14.6 & {\bf 51.5} & 9.3 & 40.6 \\

\quad \static{} & {\bf 33.5} & 48.1 & 24.7 & 61.6 & 65.9 & 67.7 & 44.7 & 75.5 & 6.1 & 23.9 & 3.7 & 22.9 \\
\quad \dynamic{} & 29.0 & 32.2 & 19.5 & 54.9& 60.1 & 61.3 & 40.7 & 73.4 & 12.6 & 21.2 & 7.4 & 29.6 \\
\quad \multipass{} & 24.3 & {\bf 52.1} & 20.6 & 57.2 & {\bf 68.5} & {\bf 70.2} & {\bf 46.0} & {\bf 76.6} & 8 & 14.3 & 5.3 & 22.0 \\
\hdashline
\quad \oracle{} & N/A & N/A & 55.9 & 82.8 & N/A & N/A & 58.5 & 84.1 & N/A & N/A & 36.5 & 68.7 \\
\bottomrule
\end{tabular}}
\caption{Results of the different methods on the \dataname{} dataset for the interpretation task and selection task evaluated on sample F1 and Exact Match (EM) by using different base LLMs from GPT and LLaMA families (LLaMA 2 (7B) for selection and CodeLLaMA 2 (7B) for interpretation).  \direct{} has the highest score on exact match followed by \joined{} for GPT-3.5 and LLaMA 2 (7B) while \multipass{} is best followed by \static{} for GPT-4. For the selection task, LLM based models are better for GPT models while LLaMA 2 struggles on this task.} 
\label{tab:main_results}
\end{table*}

\section{Experiments}
We benchmark the dataset on the above methods to explain the various challenges on the benchmark. We used GPT-3.5 (\texttt{text-davinci-003}), GPT-4 as the base LLMs from GPT family. We use LLaMA 2 (7B) for the selection task and CodeLLaMA 2 (7B) for the interpretation task from the LLaMA 2 family \citep{touvron2023llama}.

\subsection{Evaluation}
\newreplace{
For the interpretation task, we report exact match and slot-F1 of the predicted API call. 
The exact match requires 
For the selection task, we report the exact match and sentence-F1.

 For the interpretation task, the exact match requires getting every function call and its arguments equal to the ground truth. The slot-F1 is F1 score per example and then averaged over the test set.

 The order in which the API calls are generated is not important. 
 We post-processed the outputs to make punctuation consistent and lowercase both the ground truth and prediction API strings. 

For the selection task, exact match and sentence-F1 operate similarly to the above definition except that the triples are now individual standing instructions.
Similar to the API strings, all instructions are converted to lowercase. Additionally, we excluded any generated or selected instructions that were not present in the user profile. 
We provide some additional details about evaluation in Appendix~\ref{app:fine-tuning}.
}{
For both selection and interpretation tasks, we report exact match and sample F1 score. For the interpretation task, the exact match requires predicting every function call and its arguments equal to the ground truth. We treat function\_name-argument\_name-argument\_value as triples when computing F1 similar to the evaluation in dialogue state tracking \citep{dey-etal-2022-towards}. For the selection task, an exact match is when the set of predicted instructions is equal to the ground truth set of instructions.  We post-process the outputs for both the tasks (see Appendix ~\ref{app:exp}), e.g. we exclude any predicted instructions not present in the user profile.
}

\subsection{Results}

We report the results for the different methods in Table~\ref{tab:main_results}. Overall,  across all the methods, using GPT-4 as the base LLM  has better results.

Within the different ways of incorporating the selection task with the interpretation task, we find that \direct{} interpretation gives the best result (as per EM), closely followed by the \joined{} and then \static{} when using GPT-3.5 and LLaMA 2. This trend shifts for GPT-4 where \multipass{} has the best results followed by \static{} and \direct{}.
Despite the success of chain-of-thought methods in tasks like mathematical reasoning \citep{Wei2023CoT} and multi-hop question answering \citep{yoran2023answering},  we find that generating for selection and then generating API call within the same prompt
may not be suitable for incorporating standing instructions. 

\newr{We also experimented with fine-tuning smaller pre-trained models like RoBERTa \citep{liu2019roberta} and CodeT5 \citep{wang-etal-2021-codet5} for the selection and interpretation task respectively. The selection task has EM/F1 results  as 54.3/64.4. The interpretation task only reaches 7.6/37.3 suggesting that smaller models will require inclusion of techniques beyond fine-tuning such as cross-attention between the schema and the standing instructions, use of data augmentation \textit{etc}. See Appendix~\ref{app:fine-tuning} for more details.} 
\\

\noindent \textbf{Models struggle to effectively incorporate standing instructions}
The best-performing configuration across all the methods only has an exact match of 46\%. %
Considering the \oracle{} method has an exact match of 58.5\%, there is a considerable gap in performance. Incorporating standing instructions to interpret the user's context is not a trivial problem and would require approaches beyond \newr{the listed} prompting methods. Even with the gold standing instructions in \oracle{}, the models fail to achieve perfect exact match for interpretation, which shows the difficulty of the interpretation task.
We attribute this to the examples in our dataset that require understanding from different contexts - standing instructions, list of valid APIs, and the current dialogue. Further, the relevance of standing instructions can be dependent on each other. %
This may explain why we found that standard retrieval approaches fail at this task. Our findings align with the observations made in other tasks that find the retrieval of some form of context from a separate memory to be challenging \citep{weir2023ontologically,majumder2023clin}. \\ 

\noindent \textbf{Comparison across selection methods}
We find that LLM-based selection methods surpass traditional methods based on lexical statistics and embedding similarity for the GPT family as also seen in \citet{sun2023chatgpt}. Further, the gap between the \oracle{} setting in the selection module and the best-performing configuration is substantial on both exact match and F1, suggesting that selecting the relevant standing instructions explicitly from the user profile in the context of the conversation is itself challenging. This is most reflected in the LLaMA 2 (7B) results where the selection task has results worse than the BM25 and \embedding{}.

Over time, we envision the capability to add new standing instructions to user profiles, which might exceed the prompt's capacity. We anticipate that our benchmark can be useful for evaluating interesting questions in LLMs augmented with external memory \citep{lewis2020rag}. Further, decoupling the selection step would provide more transparency, as it would allow users to see their individual standing instructions that influenced the generated output \citep{liu2023trustworthy}

\subsection{Results by \scenario{}}

\begin{table}[]
\centering
\resizebox{\linewidth}{!}{

\begin{tabular}{@{}lcccccc@{}}
\toprule
Type & \oracle{} & \direct{} & \textsc{JOINT} &  \static{} & \textsc{ICL-D}  & \textsc{Multi-P} \\ \midrule
\noinstruction{} & 68.2 & 57.3 & 48.8 & 61.4 & 62.6   & 61.1\\
\simple{} & 77.9 & 67.6 & 70.5 & 69.7 & 65.0  & 70.8 \\
\chain{} & 65.5 & 56.4 & 47.3 & 59.1 & 57.9 &  60.2\\
\multi{} & 55.8 & 24.1 & 32.6 & 42.6 & 38.2 & 44.7 \\
\multidomain{} & 30.9 & 16.1 & 12.6 & 12.0 & 07.6 & 14.4\\ 
\override{} & 70.2 & 35.0 & 32.0 & 33.5   & 22.3  & 34.4 \\ \bottomrule
\end{tabular}
}
\caption{Per \scenario{} exact match on the interpretation task (GPT-4). \textsc{JOINT} is \joined{}, \textsc{ICL-D} is \dynamic{} and \textsc{MULTI-P} is \multipass{}. All the methods find \simple{} easiest while struggling at \multidomain{}. There is no consistent winning method.}
\label{tab:instr-em-g4}
\end{table}

\label{sec:scenario_results}
We break down the examples by \scenario{} in Table \ref{tab:instr-em-g4} with GPT-4 and investigate the accuracy of different methods (See Appendix \ref{app:add_res} for remaining results).
We observe that different methods display varying trends across different \scenarios{} and there is no one consistent \textit{winner} among these methods. We find that \simple{} is the easiest \scenario{} for all the methods, suggesting that LLMs do have the capacity to follow simple standing instructions. The methods perform worse on more complex \multidomain{} examples (<17\%) or \multi{} examples. These examples require sharing arguments across multiple domains, following individual standing instructions under respective domains, and reasoning across different standing instructions. 
\newreplace{From Table \ref{tab:main_results}, we note that the \multipass{} setup has an overall exact match lower than \static{}. However, the improvement in \chain{}, \multi{}, and \multidomain{} \scenario{} types over the \static{} setup suggests that another round of standing instruction selection can benefit the \scenarios{} where some complex reasoning over the instructions is required. }{Also, \multipass{} has improvement over \multidomain{} and \multi{} suggesting that another round of selection can benefit the \scenarios{} where complex reasoning over the instructions is required.}

\subsection{Qualitative Analysis}
\label{sec:error_analysis}

We annotate 100 erroneous examples each from the \direct{} and \static{} from GPT-3.5 with the most prominent error (See Table~\ref{tab:qualitative} for examples). Common errors include the hallucination of variables (Example 1) and missing arguments (Example 3) while generating API calls. For \multi{},
some predictions exclude the second API call. Further, if one of the repeating arguments has a standing instruction dependent on its value, the model does not include this conditional dependence when generating the API call (Example 2). For \multidomain{}, some predictions exclude API calls from the remaining domains (Example 3). For \direct{}, overgeneration of API calls is common. The model is likely to confuse demonstrations from \simple{} with \multidomain{}.
Another possible reason is that the model incorrectly considers many irrelevant instructions in the profile while generating the API calls. For \static{}, missing and incorrectly predicted standing instructions from the selection step produce erroneous arguments in the API calls.

\newremove{\section{Discussion}

\noindent \textbf{Identifying the subset of relevant instructions is challenging:} We compared and contrasted several baseline methods on the \dataname{} dataset. One peculiar observation is that all methods need to perform better on the selection task for our dataset especially the ones only making a single pass of selection.  The generation of the correct API call requires understanding of the user's context, the schema, the set of relevant standing instructions, as well as the dependence between standing instructions. Further, the order in which these instructions are retrieved can be important, which means the instruction selection task is no longer a retrieval of independent facts but a reasoning over attributes from the user's context. This may explain why we found that standard retrieval approaches fail at this task. Our findings align with the observations made in other tasks that find the retrieval of some form of context from a separate memory to be challenging \citep{weir2023ontologically,majumder2023clin}. 

\noindent \textbf{Impact of size of user profile:} The maximum size of the user profile in our dataset is 22 standing instructions. 
Our current methods include all the standing instructions explicitly as part of the user profile in the respective prompts. Over time, we envision the capability to add new standing instructions to user profiles, which might exceed the prompt's capacity. We anticipate that our benchmark can be useful for evaluating interesting questions in LLMs augmented with external memory \citep{lewis2020rag}. Further, decoupling the selection step would provide more transparency, as it would allow users to see their individual standing instructions that influenced the generated output \citep{liu2023trustworthy}

\noindent \textbf{Interface to incorporate standing instructions:}
Our current dataset assumes that the instructions are already provided and the user has consented to the use of the same. The \override{} \scenario{} also assumes that the user's request is preferred over standing instructions. In the future, standing instructions can be extracted from user's interactions with the system. As standing instructions become a component of a larger interface, UX design must include the user's consent to include or update such existing and inferred standing instructions. Our dataset only provides a starting point on how standing instructions can be considered by LLM-based systems.

}
\section{Related Work}

\paragraph{NL guidelines:}
\citet{gupta2022dialguide} collected and released a dataset of NL guidelines that govern the safe response generation in dialogue systems.
Compared to theirs, we showcase a more challenging retrieval setup: we have more applicable instructions on average, with rich phenomena such as \chain{} or \multi{}. %
Moreover, we are concerned with generating structured representations as a more complex final task. \citet{10.3389/frobt.2021.676814} consider a variant of standing instructions in a barista setting where the instructions consist of the favourite drink and snack of the corresponding user. Similarly, \citet{joshi2017personalization} provide a user profile consisting of age, gender, and favourite food item structured as a dictionary to \newr{ enhance the style of response generation that is appropriate to the selected attributes.
Both these works use toy scenarios \citep{Weston2015TowardsAQ}, are single-domain, contain < 5 attributes for personalisation, and use non pretrained LSTM-based sequence-to-sequence methods \citep{memory-networks} for benchmarking. Our work offers more diverse scenarios, domains (17), and attributes (150). Our instructions are more complex than maintaining user preferences in a key-value format.
}
We also explore the complexity of selecting relevant standing instructions often requiring multi-domain and multi-hop reasoning.   More recently, commercial LLM providers also offer guidelines to enhance personalisation similar to the notion of standing instructions but lacks a reported systematic evaluation (See Appendix~\ref{app:add_res}).

The use of declarative NL specifications has been explored in past work. For example,  \citet{ye2023satisfiability} use an LLM to generate a declarative task specification, coupled with an off-the-shelf automated theorem prover to derive the final answer.  \citet{weir2023ontologically} discuss methods to generate user-NPC dialogues based on game quest specifications.
Constitutional AI \cite{bai2022constitutional} identifies whether some model response violates a given rule, and then revises the response accordingly. 

Closely related to the use of standing instructions is also learning from feedback \citep{labutov-etal-2018-learning, tandon-etal-2022-learning, madaan2022memory}, where the goal is to maintain a memory of user-provided feedback and use it to augment the knowledge used by question-answering models at test time. Analogously, standing instructions can also be seen as a form of memory.

\paragraph{Personalisation:} Personalisation in dialogue has been extensively studied (\citet{li-etal-2016-persona, zhang-etal-2018-personalizing, majumder2020like}; \textit{inter-alia}) where the personality traits are provided through NL statements. However, all these works focus on providing a persona to the bot to generate more engaging responses rather than assisting the users in completing their request. 

In a broader sense, learning from preferences has been fundamental to improving user experience. These include 
personalised review generation \citep{li2020knowledge}, personalised search results through collaborative filtering \citep{micarelli2007personalized} or leveraging a profile of user interests \citep{Speretta2005PersonalizedSB}. 
\citet{salemi2023lamp} explored personalised text generation with LLMs on tasks such as article generation given past articles authored by the user. Our work provides incorporation of preferences explicitly through standing instructions allowing better understanding of the generated result.

\section{Conclusion}
We proposed the use of standing instructions - a set of natural language statements that contain the user's preferences to enhance the interpretation of the user's requests. To facilitate this, we created \dataname{}, a language-to-program dataset based on SGD. This enabled us to explore two tasks: standing instruction selection and interpretation task of generating API calls which are conditioned on the selected instructions and conversational context.
We experimented with several methods for the selection and interpretation tasks. Our results show that while LLMs are somewhat capable of incorporating standing instructions as an additional context, their usage of standing instructions is far from perfect. The models struggled to select the instructions in the user profile that were relevant to the given dialogue, which in turn affected the interpretation task. Moreover, as \scenarios{} become more intricate and involve complex reasoning or interactions among the respective standing instructions, the interpretation of these instructions becomes increasingly challenging for the methods. This calls for the development of new approaches in incorporating standing instructions, reasoning-based retrieval, and memory-augmented representations.

\section{Ethics Statement}
Our dataset is based on SGD \citep{rastogi2020towards} which consists of fictional conversations. The real world named entities such as restaurant names for the dataset were sampled from Freebase while date/times were sampled synthetically. No human names or any personal information is present in the dataset. Our task involves API call generation in a constratined setup which generally does not produce harmful or toxic responses.

\section{Limitations}
Our task setup is limited to generating API calls for the current turn. In an ideal scenario, the LLM or the service should also display the results in a user-friendly format, like natural language or Mark-down, and perhaps confirm with the user before executing the call. Our dataset is not accompanied by the results from respective API calls or replies from the system due to the unavailability of results from the base dataset. The different \scenarios{} in our dataset are not exhaustive and future work could look into expanding them. \newr{The number of APIs in the dataset is 17 that currently fits in the prompt. In future iterations, as the number of APIs will increase beyond the prompt length, we would need to incorporate techniques from \citet{qin2023toolllm,ye2024rotbench} as an additional step to select the right APIs.}

As our dataset is derived from an existing academic task-oriented dialogue dataset, it is useful for testing methods, but we caution readers that \newr{real-world services will include more complex standing instructions, domains, and user scenarios. The standing instructions were derived from templates and then adequately paraphrased. Despite this, we find it to be a challenging and non-trivial benchmark as evident in our results section}  
Further, preferences stated explicitly by a human user would likely take a wider range of natural language forms.  Preferences deduced from the user's past history might take a non-linguistic form, as in recommendation systems; they might be uncertain or soft constraints that cannot be passed directly as arguments to simple search APIs.

\bibliography{custom,anthology}
\bibliographystyle{acl_natbib}
\appendix
\section{Dataset Construction Details}
\label{app:data}
We provide further details about dataset construction. 

\paragraph{Forming examples for different \scenario{}s:} We do not need to extract any standing instructions $z$ for examples in \noinstruction{}. For examples in \simple{}, each (domain, slot, value) triple was extracted and written in natural language via an if-then template and appropriately paraphrased. Since each slot is independent of each other, this set of instructions form $z$. \chain{} examples were formed by creating a hierarchy of slots associated with the same domain like \textit{seating\_class} is dependent on \textit{airlines}. If the subsequent dialogue states contained the same dependent slots, then that example was categorized as a \chain{} example, where the primary slot value was obtained from the dialogue or one of the standing instructions. 
\multidomain{} examples were dialogues from SGD that were inherently multi-domain because they required API calls from different domains. These \scenario{}s were created through a deterministic process based on the existing SGD data. 

\multi{} examples were formed by duplicating one of the ground truth standing instructions from \simple{}, \chain{} and \multidomain{}, and substituting an argument value with another relevant entity. Meanwhile, \override{} examples were formed with examples from \simple{} or \chain{}. We added information that conflicts with the gold standing instruction like asking for \textit{Mexican} restaurants when the standing instruction is about preference for \textit{Italian} restaurants. \newr{We provide examples for the remaining \scenario{}s in Table~\ref{tab:rem_examples}.}

\paragraph{Sampling instructions for user profile}:We drew $M$ instructions uniformly from the range $[3,12]$. In particular, we drew the distractor instructions before splitting the dataset into train/dev/test, so training examples were constructed with some distractors sourced from the test set.  
Given this dataset, however, our experiments followed the usual protocol of holding out the test set while constructing our systems.

\paragraph{Post-processing: }
\new{We also included several rounds of post-processing on the dataset to remove undesirable or unrealistic situations that arise either through the noise in the base dataset or our extraction process. We removed examples with domain mismatches in case of \multidomain{} such as requesting music which is followed by a request for bus ticket booking.}
We unified domains such as \textit{Restaurant\_1}, \textit{Restaurant\_3} as \textit{Restaurants}. \textit{Restaurant\_2} was renamed as \textit{HouseStays}. We also deduplicated the slot names under these domains like \textit{location} and \textit{area} was converted to \textit{area}. Similarly, the \textit{Services} domain was expanded as \textit{Salons}, \textit{Doctors}, and \textit{Dentists} instead. All the examples were constructed only from the domains and examples available in the training set of SGD.
In addition to removing domains whose combination doesn't make sense in the \multidomain{} \scenario{}, we also remove \multidomain{} examples which do not have any attributes for the second domain. 

The instructions obtained through the above deterministic process were templated. For paraphrasing the templated instructions, we prompted GPT-3 to generate paraphrases with three distinct prompts to promote diversity. \\
{\small
\noindent Prompt 1: \texttt{Write a colloquial paraphrase for the given sentences. Refrain from using if then format} \\
\noindent Prompt 2: \texttt{Reword the following in your own words. Keep the same meaning. Change the sentence structure to exclude if then format:   }\\
\noindent Prompt 3: \texttt{Reword the following in your own words. Keep the same meaning.  Make the sentences sound like instructions or commands. \newline
Change the sentence structure to exclude if-then format. If the sentence starts with ``If I ask for xyz'', also reword that xyz part.} }\\
We replace the templated standing instruction randomly with one of the paraphrases leading to 4097 unique instructions across the dataset.

\newcolumntype{K}{>{\arraybackslash}m{1.5cm}}
\newcolumntype{L}{>{\arraybackslash}m{3.9cm}}
\newcolumntype{M}{>{\arraybackslash}m{4.2cm}}

\begin{table*}[]
\scriptsize
\begin{tabular}{KMMM}
\toprule

 & \override{} & \noinstruction{} & \multidomain{} \\ \midrule

 User Profile \newline($u$) & >When I request Restaurants, I prefer Italian cuisine.\newline >If I'm looking for a doctor, I'd rather have a General Practitioner.
\newline >If I'm opening a bank account, I want it to be a savings account.\newline >I'd like to get a Doctor in San Rafael if I can. \newline \dots & >Request Restaurants with Filipino cuisine as my preference.\newline >Request Music by Iggy Azalea as my preferred artist.\newline >If I'm looking to go to the movies, my go-to theatre is Airport Stadium Cinemas.\newline  >If I'm looking for a flight, my go-to airline is Alaska Airlines.\newline >Request Events, specifically Sports events. & >When I request Movies, I typically enjoy ones that are comedic.\newline >My first choice when requesting Travel is Vegas \newline >When it comes to Hotels, I prefer ones that are rated 1-star.\newline >My go-to theater for Movies is AMC Bay Street.\newline >If I'm looking into Travel, I should also check out Hotels\newline >I'd like my travel to be kid-friendly. \newline \dots \\

Relevant\newline Standing\newline Instructions ($z$) & >\textcolor{red}{I'd like to get a Doctor in San Rafael if I can.} & None & >\textcolor{red}{My first choice when requesting Travel is Vegas}
\newline >\textcolor{red}{If I'm looking into Travel, I should also check out Hotels.} %
\newline >\textcolor{red}{When it comes to Hotels, I prefer ones that are rated 1-star.} \newline \textcolor{red}{I'd like my travel to be kid-friendly.}  \\\\

Conversation ($x$) &  \textcolor{blue}{\emph{User}: I need to find a Gynecologist} & \textcolor{blue}{\emph{User}: Can you help me find some attractions to see?} \newline \emph{Agent:} Where should I look? \newline \textcolor{blue}{\emph{User}: How about in KL?}  & \textcolor{blue}{\emph{User}: User: Any good tourist traps out there?} \\\\

API calls ($y$) & 
\begin{lstlisting}[basicstyle=\scriptsize\ttfamily, breaklines=true]
GetDoctors(
  type="Gynecologist", 
  location="San Rafael")
\end{lstlisting}
&
\begin{lstlisting}[basicstyle=\scriptsize\ttfamily, breaklines=true]
GetTravel(
  location="KL")
\end{lstlisting}
&
\begin{lstlisting}[basicstyle=\scriptsize\ttfamily, breaklines=true]
GetTravel(
  good_for_kids="True"
  location="Vegas")
GetHotels(
  average_rating="1", 
  location="Vegas")
\end{lstlisting}

\\

\bottomrule
\end{tabular}
\caption{Some examples from \dataname{}. 
(1) In \override{}, user requests for an attribute that is against the standing instructions (``Gynecologist'' v/s ``General Practionier''). (2) In \noinstruction{}, the user makes a request which is not affected by the standing instructions. (3) In \multidomain{}, the examples contain an instruction which requires invoking a hotel search for the same location when user requests for places to visit.
}
\label{tab:rem_examples}
\end{table*}

\section{Experiment Details}
\label{app:exp}
\subsection{Setup}
For the selection experiments involving BM25 and Contriever, $N$ was varied from 1 to 10 and chosen according to the best exact match on the dev set ($N$=4 for BM25, $N$=2 for \embedding{}). 
For LLMs, the $K$ for demonstrations was varied among \{3,5,8\}, with $K$=5 being best for \dynamic{} and other interpretation tasks. For the \multipass{} experiments, we varied $K$ for three additional rounds and found that providing one additional pass had the best results on the development set. %
We use temperature of 0 while decoding from the LLMs unless specified otherwise. %
We will provide the prompt templates for the different experiments. %
We use LLaMA 2 7B\footnote{\url{https://huggingface.co/meta-llama/Llama-2-7b-hf}} for the selection experiments. As our API calls are similar to the python syntax of a function, we use CodeLLamA 2 7B (instruction fine-tuned) \footnote{\url{https://huggingface.co/codellama/CodeLlama-7b-Instruct-hf}} for the interpretation experiments. We also found CodeLLaMA 2 (7B)'s results to be better than LLaMA 2 (7B) for the interpretation task on the validation set.
We use 2 24GB GPUs, batch size of 1, full precision models for the these experiments. It takes approx 48 hours to make a pass over the entire test set.

For evaluation, all the outputs were converted to lowercase and double quotes were unified to a fixed unicode. Using ``vs'' and ``versus'' was unified to ``versus''. The models were not penalised if they produced \textit{subcategory} instead of \textit{event\_type} arising due to the noise in the base dataset. For the interpretation evaluation, the API calls were converted to function\_name-slot-value triples per slot-value per API call. In the case of examples multiple API calls, the models had a tendency to include every attribute in a single API call instead of separate API calls. To penalise this in the exact match, if the number of predicted API calls was not equal to the number of ground truth API calls the model received an exact match of 0. 

\newr{\subsection{Prompts}}
We shall now list the prompts used in our experiments.
\subsubsection{Selection Task}
\begin{figure*}[htb]
{\tiny
   \begin{Verbatim}[commandchars=+\[\]]
Standing instructions allow a user to add preferences or requirements that an agent would like to consider when generating its responses. 
The user's current utterance in the dialogue has priority over standing instructions. 
For the given dialogue, which of the following standing instructions are applicable? If no standing instructions are applicable, then generate "None". 

Standing Instructions:
<demonstration standing instructions>

Dialogue:
<demonstration dialogue>

Applicable Standing Instructions:
<demonstration applicable standing instructions>
<EOS>

Standing Instructions:
<test standing instructions>

Dialogue:
<test dialogue>

\end{Verbatim}
}
\caption{Prompt for the ICL Selection task. The number of examples and the type of examples will vary according to the experiment}
\label{fig:static-prompt}
\end{figure*}
\begin{figure*}[htb]
{\tiny
   \begin{Verbatim}[commandchars=+\[\]]
You are designing a parser that takes in a user utterance and some standing instructions and outputs a set of API calls. 
Every API call consist of "GetX" where X is domain name and uses slot names listed below as arguments.  
We list the domain name followed by the list of possible slot names. Some slot names can be categorical or boolean
The values for the arguments can come from the user's dialogue or standing instructions. If the user requests a slot name and no value is found, use "?". 
If the user requests dontcare, use value as "any".
Standing instructions allow you to add preferences or requirements that you’d like to consider when generating the parser. 
If standing instructions are applicable across multiple domains, place an API call per situation per domain. 
If some of the applicable standing instructions have instructions of similar type, place multiple API calls respecting the standing instructions.
If some slots are applicable across several domains, generate the respective slot names for the respective domains.

Schema:
Banks: recipient_account_name, amount, recipient_account_type
Buses: origin, departure_date, fare_type, transfers, price, group_size, destination, destination_station_name, origin_station_name, departure_time
Events: event_name, city, category, event_location, number_of_tickets, time, address_of_location, date, venue_address, event_type
Flights: origin, inbound_arrival_time, is_redeye, outbound_departure_time, outbound_arrival_time, inbound_departure_time, return_date, airlines, 
seating_class, refundable, number_stops, destination_airport, departure_date, fare, destination, passengers, origin_airport
Homes: pets_allowed, visit_date, address, property_name, rent, number_of_baths, area, number_of_beds, furnished, phone_number
Hotels: has_wifi, average_rating, check_out_date, price, pets_welcome, number_of_days, location, check_in_date, phone_number, 
number_of_rooms, street_address, hotel_name
HouseStays: rating, phone_number, has_laundry_service, check_out_date, total_price, check_in_date, address, number_of_adults, where_to
Media: title, directed_by, subtitles, genre
Movies: theater_name, movie_name, price, show_date, location, show_time, number_of_tickets, genre, show_type, street_address
Music: song_name, year, album, artist, genre, playback_device
RentalCars: dropoff_date, pickup_time, pickup_city, pickup_date, total_price, car_type, car_name, pickup_location
Restaurants: price_range, restaurant_name, city, has_live_music, serves_alcohol, time, date, phone_number, cuisine, street_address, party_size
Salons: is_unisex, average_rating, city, appointment_date, appointment_time, stylist_name, phone_number, street_address
Dentists: dentist_name, phone_number, offers_cosmetic_services, city, appointment_date, appointment_time, address
Doctors: doctor_name, city, average_rating, appointment_date, appointment_time, type, phone_number, street_address
Travel: good_for_kids, category, attraction_name, location, phone_number, free_entry
Weather: city, temperature, date, precipitation, humidity, wind


Further, following slots have categorical values:
recipient_account_type: checking, savings
fare_type: Economy, Economy extra, Flexible
(Events) category: Place of Worship, Theme Park, Museum, Historical Landmark, Park, Tourist Attraction, Sports Venue, Shopping Area, 
Performing Arts Venue, Nature Preserve
event_type: Music, Sports
seating_class: Economy, Premium Economy, Business, First Class
refundable: True, False
airlines: United Airlines, American Airlines, Delta Airlines, Southwest Airlines, Alaska Airlines, British Airways, Air Canada, Air France
show_type: regular, 3d, imax
playback_device: TV, kitchen speaker, bedroom speaker
(Doctors) type: Gynecologist, ENT Specialist, Ophthalmologist, General Practitioner, Dermatologist
car_type: Compact, Standard, Full-size
price_range: inexpensive, moderate, expensive, very expensive


Further, following slots are boolean:
has_wifi, pets_allowed, subtitles, offers_cosmetic_services, has_laundry_service, is_unisex,

good_for_kids, has_live_music, pets_welcome, serves_alcohol, is_redeye, furnished, free_entry


Dialogue:
<demonstration dialogue>

Standing Instructions:
<demonstration instructions>

API Calls:
<demonstration api calls>
<EOS>

Dialogue:
<test dialogue>

Standing Instructions:
<test instructions>

API Calls: 
    \end{Verbatim}
}
\caption{Prompt used for interpretation experiments. We include the template for demonstration examples and test examples in this figure. Note the demonstration examples will be repeated based on the number of demonstration examples used}
\label{fig:parsing_general}
\end{figure*}
For the selection tasks, the prompt is described in Fig~\ref{fig:static-prompt}. For the \multipass{} experiments, an additional instruction was added to the prompt ``If some instructions are missing from the current set, generate those instructions under Remaining Applicable Standing Instructions''. The test example consists of ``Applicable Standing Instructions'' from the previous iteration and ``Remaining Applicable Standing Instructions'' is appended with every demonstration. 

\raggedbottom
\subsubsection{Interpretation Task}
We describe the prompt in Fig~\ref{fig:parsing_general} used for Direct Interpretation and \textsc{Selection-Then-Interpretation} methods. The set of standing instructions will vary depending on the type of experiment. For \textsc{Joint Selection and Interpretation}, the prompt includes an additional sentence ``For the following dialogue, first generate the appropriate applicable standing instructions from the user profile and then generate API calls based on the dialogue and the selected standing instructions.'' between ``Standing instructions allow you to add preferences or requirements that you’d like to consider when generating the parser.'' and ``If standing instructions are applicable across multiple domains, place an API call per situation per domain''. The demonstration and test example format look as Fig~\ref{fig:cot-example}.
\begin{figure}[H]
{\tiny
   \begin{Verbatim}[commandchars=+\[\]]
Dialogue:
<demonstration dialogue>

User Profile:
<demonstration standing instructions>

Applicable Standing Instructions
<applicable demonstration standing instructions>

API Calls:
<demonstration api calls>
<EOS>

Dialogue:
<test dialogue>

User Profile:
<test standing instructions>
 \end{Verbatim}
}
\caption{Demonstration and test example format for Select-And-Interpret experiments}

\label{fig:cot-example}
\end{figure}

\section{Additional Results}
\label{app:add_res}
\newr{
\subsection{Dependence on paraphrasing}
We experiment with five different random seeds for the dataset creation, creating five different versions of the dataset. We evaluate the \direct{} method on the \textsc{LLaMA-2} model for the development set. The average exact match across these datasets is 15.1$\pm$0.7 suggesting only small variance. 

\subsection{Fine-tuning experiments}
\label{app:fine-tuning}
We fine-tune smaller pre-trained models to benchmark them on the \dataname{} dataset. 
\begin{table}[]
\small
\centering
\begin{tabular}{@{}llll@{}}
\toprule
\begin{tabular}[c]{@{}l@{}}Selection\\ Method\end{tabular} & \begin{tabular}[c]{@{}l@{}}Interpretation\\ Training Data\end{tabular} & EM & F1 \\ \midrule
QA & User Profile & 11.2 & 43.0 \\
QA & Applicable & 12.2 & 42.4 \\
Oracle & User Profile & 13.2 & 47.3 \\
Oracle & Applicable & 15.5 & 50.1 \\ \bottomrule
\end{tabular}
\caption{Interpretation task scores when fine-tuned with User Profile and Applicable standing instructions respectively for the interpretation task while using ``Oracle'' or standing instructions obtained from a fine-tuned QA model (based on RoBERTa). Fine-tuned models struggle at the interpretation task and a model trained with applicable standing instructions is better.}
\label{tab:fine-tuning}
\end{table}

\noindent \textbf{Selection Task}: We start with trained extractive question-answering system that uses RoBERTa-base \citep{liu2019roberta} as the encoder and SQuAD 2.0 \citep{rajpurkar-etal-2018-know} as the training dataset. \footnote{https://huggingface.co/deepset/roberta-base-squad2} 
In our setup, the dialogue forms the paragraph, and [``yes''] and [``no''] are appended to the start of the dialogue. The question is ``Is the standing instruction X applicable" and if the predicted answer is ``yes'', the respective instruction X is selected. This process is repeated for every instruction in the user profile. We further fine-tune this question-answering model by converting every example in the training set into such a format.

\noindent \textbf{Interpretation Task}: We fine-tune a code-specific pre-trained model, namely CodeT5 (220 M) \citep{wang-etal-2021-codet5} on \dataname{} dataset. As this is a Sequence-to-Sequence model, the input consists of the dialogue concatenated with the instructions from the user profile and the output consists of the API calls. This is similar to the \direct{} method discussed in Section~\ref{sec:methods}. To simulate the \textsc{Select-Then-Interpret} paradigm, we design two interpretation models, one using all the standing instructions from the user profile and the other using the applicable standing instructions only (Applicable).

\noindent \textbf{Results}: The stand-alone selection task leads to an Exact Match/F1 score of 54.3/64.4 which provides a strong baseline result. The \direct{} interpretation results in  7.6/37.3 indicative of a need for better interpretation models. The results for \textsc{Select-Then-Interpret} with smaller models are reported in Table~\ref{tab:fine-tuning}. We find that \textsc{Select-Then-Interpret} has improved results over  \direct{} unlike some of the LLM results. We further find that using applicable standing instructions during the training of the interpreter leads to better results. Lastly, we find that even with oracle instructions and interpreter trained with applicable instructions, the interpretation task has poor capabilities.

}

\subsection{Scenario Type results for GPT-3.5 and LLaMA 2}

\begin{table}[]
\centering
\resizebox{\linewidth}{!}{
\begin{tabular}{@{}lcccccc@{}}
\toprule
Type & \oracle{} & \direct{} & \textsc{JOINT} & \textsc{ICL-D} & \static{} & \textsc{MULTI-P}\\ \midrule
\noinstruction{} & 65.3 & 45.9 & 37.9 & 54.4 & 58.5 & 29.4 \\
\simple{} & 80.3 & 56.2 & 56.5 & 41.8 & 28.5 & 36.5 \\
\chain{} & 65.3 & 41.8 & 34.1 & 27.6 & 19.1 & 34.1 \\
\multi{} & 40.0 & 11.5 & 11.5 & 8.8 & 4.1 & 9.7 \\
\multidomain{} & 23.2 & 3.5 & 3.2 & 0.6 & 0.3 & 1.2\\ 
\override{} & 70.3 & 34.1 & 26.2 & 17.1 & 6.8 & 14.7 \\ \bottomrule
\end{tabular}
}
\caption{Per \scenario{} exact match on the interpretation task (GPT-3.5). \textsc{ICL-D} is \dynamic{} and \textsc{MULTI-P} is \multipass{}. All the methods find \simple{} easiest and struggle on \multidomain{}. Different methods show different trends without a consistent winner.  }
\label{tab:instr-em}
\end{table}

\begin{table}[]
\centering
\resizebox{\linewidth}{!}{

\begin{tabular}{@{}lcccccc@{}}
\toprule
Type & \oracle{} & \direct{} & \textsc{JOINT} &  \static{} & \textsc{ICL-D}  & \textsc{Multi-P} \\ \midrule
\noinstruction{}   & 45   & 24.4 & 23.8 & 4.1 & 27.9 & 17.6  \\
\simple{}          & 62.1 & 36.2 & 37.1 & 8.8 & 7.4  & 5.3  \\
\chain{}         & 48.2 & 17.1 & 17.4 & 1.5 & 1.5  & 2.9  \\
\multi{} & 19.4 & 5.3  & 4.4  & 0.9 & 1.5  & 0.6  \\
\multidomain{}       & 3.2  & 1.2  & 0.6  & 0.3 & 0.3  & 0.0  \\
\override{}               & 48.8 & 8.2  & 7.4  & 7.4 & 6.5  & 5.8  \\ \bottomrule
\end{tabular}
}
\caption{Per \scenario{} exact match on the interpretation task (LLaMA 2). \textsc{JOINT} is \joined{}, \textsc{ICL-D} is \dynamic{} and \textsc{MULTI-P} is \multipass{}. All the methods find \simple{} easiest while struggling at \multidomain{}. There is no consistent winning method.}
\label{tab:instr-em-llama}
\end{table}

We report the results by \scenario{} for experiments using base LLM as GPT-3.5 in Table \ref{tab:instr-em} and LLaMA 2 in Table \ref{tab:instr-em-llama}. The trends are similar to the trends discussed in Section~\ref{sec:scenario_results}.

\subsection{OpenAI's Custom Instructions}
OpenAI also recently reported the introduction of custom instructions\footnote{\url{https://openai.com/blog/custom-instructions-for-chatgpt}} that allow the users to add requirements or preferences that ChatGPT should consider when generating the responses. This is similar to our notion of standing instructions. To test the effectiveness of this feature (free version), we use the instructions from the user profile as ``custom instructions''. We pose the API generation task as a standalone task and hope for the model to directly incorporate the standing instructions from the custom instructions. We also use the \textbf{ICL} setup to provide examples about the task as discussed in Section \ref{sec:select-then-interpret}. As this effort requires manual copy-pasting of examples, 
we randomly selected and evaluated 17 examples per type, amounting to 102 test examples. While not directly comparable with Table~\ref{tab:main_results}, the exact match for the interpretation task on this subset is 15.6 and the slot F1 score is 45.5. Thus, the model does not necessarily incorporate the correct custom instructions every time. It is prone to copying arguments from the demonstration example as well as hallucinating the arguments and their values. For some examples, the model is prone to over-generation of API calls and other unrelated text.
We remark that due to the opacity of the ``custom instructions'' UI, we do not know the prompt or exact LLM API call and we have observed nondeterminism in the outputs.

\subsection{Qualtitative Analysis}
We list some examples exhibiting prominent errors discussed in Section ~\ref{sec:error_analysis} in Table~\ref{tab:qualitative}. 
\newcolumntype{L}{>{\arraybackslash}m{3.2cm}}
\newcolumntype{K}{>{\arraybackslash}m{2.2cm}}
\newcolumntype{P}{>{\arraybackslash}m{1cm}}

\begin{table*}[]
\scriptsize
\resizebox{\textwidth}{!}{%
\begin{tabular}{PKKLLP}
\toprule
\scenarioCap{} & Conversation & Gold SI & Gold API & Predicted API & Error \\ \midrule
\simple{} & \textit{User}: I am looking for a place of interest to visit. & >If I'm looking to go somewhere, my top choice is New York City. & \texttt{GetTravel(category="Tourist Attraction", location="New York City)} & \texttt{GetTravel(\textcolor{red}{category="Place of Interest"}, location="New York City")\newline \textcolor{red}{GetMusic(genre="Country")}\newline \textcolor{red}{GetMedia(genre="Drama", directed\_by="Qui Sheng")}}\newline $\dots$ & Hallucination \\
\textsc{Multi Preference} & \textit{User}: I fancy watching a movie at home this evening & >When I request Media, my favorite type is adventure. \newline \textit{>If you're looking for an adventure movie, search for one directed by Joel Coen}. \newline >When I request Media, my favorite type is biographical. & \texttt{GetMedia(\textcolor{red}{directed\_by="Joel Coen"}, genre="adventure")\newline GetMedia(genre=\newline"biographical")} & \texttt{GetMedia(genre="adventure"), \newline GetMedia(genre=\newline"biographical")} & Skipped Instruction \\\\
\textsc{Multi Domain} & \textit{User}: I'm looking for some interesting attractions to visit, specifically a Museum. Can you help me find any? & >I prefer London as my destination when I travel. \newline >If I'm looking into Travel, I should also check out Hotels. \newline >I prefer to stay in hotels that have a two-star rating when I am making reservations. & 
\texttt{GetTravel(category="Museum", \textcolor{red}{location="London"}) \newline \textcolor{red}{GetHotels(average\_rating="2", location="London")}}
& \texttt{GetTravel(category="Museum")} & Missing API \newline Prediction \\ \bottomrule
\end{tabular}}
\caption{Examples of prominent errors across the \direct{} and \static{} methods (with GPT-3).  
The incorrectly predicted or missing arguments and function calls are marked in \textcolor{red}{red}. The \direct{} models tend to produce several unrelated API calls (first example). Both \direct{} and \static{} have a tendency to miss an argument that is only dependent on one of the attributes in \multi{}, in this case missing the director \textit{Joel Coen}. Majority of predictions in \multidomain{} fail at generating the API calls for the second domain.} %
\label{tab:qualitative}
\end{table*}

\end{document}